# Diagnosis of aerospace structure defects by a HPC implemented soft computing algorithm

Gianni D'Angelo, Salvatore Rampone
University of Sannio
Dept. of Science and Technology
Benevento, Italy
{dangelo, rampone}@unisannio.it

*Abstract*— This study concerns with the diagnosis of aerospace structure defects by applying a HPC parallel implementation of a novel learning algorithm, named U-BRAIN. The Soft Computing approach allows advanced multi-parameter data processing in composite materials testing. The HPC parallel implementation overcomes the limits due to the great amount of data and the complexity of data processing. Our experimental results illustrate the effectiveness of the U-BRAIN parallel implementation as defect classifier in aerospace structures. The resulting system is implemented on a Linux-based cluster with multi-core architecture.

*Keywords*— *Non-destructive testing; learning algorithm; parallel computing; HPC; signature-based classifier; eddy current.*

## I. INTRODUCTION

The use of composite materials in the aerospace industry is growing rapidly, especially in the production of the components the use of which sees them subjected to heavy loads and efforts. Due to their unique mechanical properties, namely, high strength-to-weight ratio, high fracture toughness, and excellent corrosion resistance properties, they are used at critical points in the construction of an aircraft [1,2]. They are widely used in the outer covering of the aircraft, such as flaps, hatches, sides of the engine, floors, rudders, elevators, ailerons etc. The composite material design and manufacturing technologies have matured to a level that Boeing Company is using composite material for 50% of the primary structure in its 787 program. Unfortunately, there is a great variety of possible manufacturing defects that regards those materials [3]. The most widespread types of defects are the following:

- Delamination between plies of outer skin, parallel to surface;
- Matrix crack;
- Disbanding between the outer skin and the honeycomb core;
- Fiber fracture;
- Cracked honeycomb core parallel to the inspection surface;
- Crushed honeycomb core in parallel to the area;
- Disbonding between inner skin and honeycomb core;
- Fluid ingress in honeycomb core.
- Damages induced by the stress, environment influences and others.
- Wear, scratch, indentation and cleft
- Creep deformation.

These defects are difficult to diagnose and analysis is strongly influenced by many factors that may also arise from the complexity of manufacturing processes. In addition, some techniques of inspection and/or some detection equipment may have systematic errors or accidental ones. The presence of defects and damages pose a significant threat to the safety of composite structures. Composite materials are mostly used in aerospace structures, and their structural reliability and safety is particularly critical.

Non-Destructive Testing (NDT) allows one to implement a control over the material at different stages of its evolution and permits to safeguard the integrity of the structure during the analysis. Visual and strike method, optical holography, X-ray, ultrasonic wave, eddy current testing and infrared detection, X-ray and ultrasonic C-scan are the most methods used. Due to the heterogeneity of the composite structure, the Non-Destructive Testing of composites are very complex and sometimes several methods will take to test the same component [4]. For this reason, the accuracy of diagnosis of composite materials is determined not only by physical methods to obtain experimental data but also with mathematical models and advanced methods of data processing [5]. The analysis of the data, generally obtained from tests based on multi-parameter control, is one of the possible ways to increase the effectiveness and reliability of non-destructive testing of composites. The methods of spectrum analysis and pattern recognition are often used in multi-parametric control for data processing [6]. However, the application of these methods requires sophisticated techniques for processing signals that lead to the solution of nonlinear equations complex with a high number of variables [7]. The difficult and sometimes impossible solution of these equations lead to a reduction in the efficiency of the system of non-destructive testing. These difficulties also do not allow the automation of the test and deprive their of the same dynamism typical of a system able to adapt to changes in the parameters of the testing system at run-time. Non-destructive testing of composites should be performed with methods able to collect the most comprehensive information about new defects, expand existed base of defects and increase diagnostics system precision in runtime. Furthermore data processing in defects diagnosis has to deal with great amount of data and numerous elements are processed with the same operation.

An alternative method of data processing and construction of decision rules for multi-parameter non-destructive testing of composite materials is to use Soft Computing techniques [8]. Soft Computing methods as neural networks [9] are proven to be effective in non destructive testing. In recent years Support vector machines (SVMs) show comparable or better results than neural networks and other statistical models [10], and they are mostly used to classify the defects [11].

This work concerns with the diagnosis of aerospace structure defects by applying a HPC parallel implementation of a novel learning algorithm, named U-BRAIN. The Soft Computing approach allows advanced multi-parameter data processing in composite materials testing. The HPC parallel implementation overcomes the limits due to the great amount of data and the complexity of data processing. The system has been tested on the automated classification of eddy current signatures. Eddy current testing is one of the most extensively used non-destructive techniques for electrically inspecting materials at very high speeds that does not require any contact between the test piece and the sensor [12]. Our experimental results illustrate the effectiveness of the U-BRAIN parallel implementation as defect classifier in aerospace structures. The resulting system is implemented on a Linux-based cluster with multi-core architecture.

The paper is organized as follows. In Section 2 we describe the U-BRAIN algorithm and its HPC implementation. A 'case of study' NDT application is reported in Section 3. Section 4 is devoted to the conclusions.

## II. U-BRAIN Algorithm and HPC implementation

The U-BRAIN (Uncertainty-managing Batch Relevance-based Artificial INtelligence) algorithm [13] is a learning algorithm that finds a rule described as a Boolean formula (f) in disjunctive normal form (DNF) [14], of approximately minimum complexity, that is consistent with a set of data (instances). The conjunctive terms of the formula are computed in an iterative way by identifying, from the given data, a family of sets of conditions that must be satisfied by all the positive instances and violated by all the negative ones; such conditions allow the computation of a set of coefficients (relevances) for each attribute (literal), that form a probability distribution, allowing the selection of the term literals. This algorithm was originally conceived for recognizing splice junctions in human DNA [15-16]. Splice junctions are points on a DNA sequence at which "superfluous" DNA is removed during the process of protein synthesis in higher organisms. The general method used in the algorithm is related to the STAR technique of Michalski [17], to the candidate-elimination method introduced by Mitchell [18], and to the work of Haussler [19]. The algorithm was then extended by using fuzzy sets [20], in order to infer a DNF formula that is consistent with a given set of data which may have missing bits. The great versatility that characterizes it, makes U-BRAIN potentially applicable in every industry and science in which there is data to be analyzed, such as the financial world, the aviation industry, the biomedical field. UBRAIN models a process starting from a limited number of features of interest from examples, data structures or sensors. A scheme of the U-BRAIN algorithm is reported in the Appendix.

However, according to the Landau's symbol [21] to describe the upper bound complexity with big O notation, the overall algorithm time complexity is $\approx O(n^5)$ and the space complexity is in the order of $\approx O(n^3)$. In order to overcome the limitations related to high computational complexity, recently an high performance parallel based implementation of U-BRAIN has been realized [22]. Mathematical and programming solutions able to effectively implement the algorithm U-BRAIN on parallel computers have been found; a Dynamic Programming model [23] has been adopted. Finally, in order to reduce the communication costs between different memories and, then, to achieve efficient I/O performance, a mass storage structure has been designed to access its data with a high degree of temporal and spatial locality [24]. Then a parallel implementation of the algorithm has been developed by a Single Program Multiple Data (SPMD) technique together to a Message-Passing Programming paradigm. In Fig. 1 and Fig. 2 are depicted the speed-up of the parallel implementation varying the number of processors on two standard datasets (HS3D and COSMIC) [25, 26]

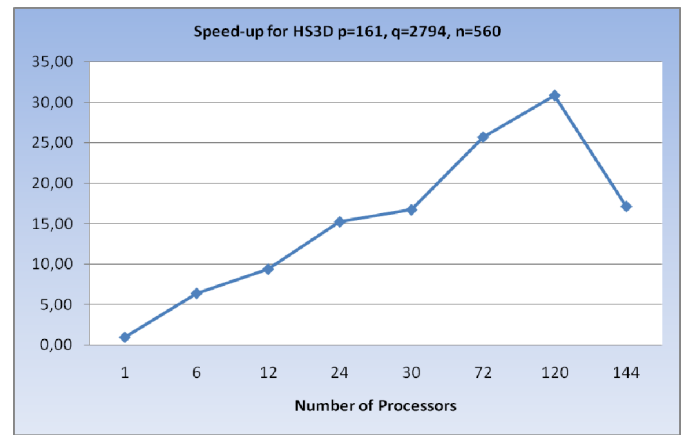

Fig. 1. Speed-up of the U-BRAIN parallel implementation on HS3D varying the processor number.

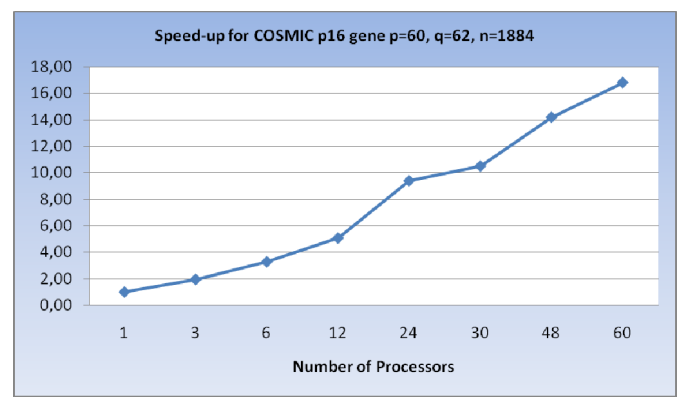

Fig. 2. Speed-up of the U-BRAIN parallel implementation on COSMIC p16 gene varying the processor number.

TABLE I. U-BRAIN RESULTS

| Test | Ten fold cross validation results | | |
|---|---|---|---|
| | Rule | Training Error | Validation Error |
| 1 | x36 x37 x_46 + x26 x_30 + x_15 x68 x_90 + x8 x_44 + x50 x_87 | 0.00 | 0.00 |
| 2 | x_15 x68 x_71 x83 + x_46 x78 x80 + x_35 x37 x67 + x_11 x14 x37 + x_6 x36 x91 + x_41 | 0.00 | 0.00 |
| 3 | x36 x37 x_46 + x_22 x37 x58 + x28 x_51 x_90 + x_41 x_44 + x7 x_35 x50 | 0.00 | 0.00 |
| 4 | x_15 x_51 x68 + x36 x37 x_46 + x_41 x_63 + x32 x_44 + x5 x_38 | 0.00 | 0.00 |
| 5 | x_15 x68 x_71 x83 + x36 x_46 x66 x100 + x37 x68 x_70 + x32 x_44 + x5 x_96 | 0.00 | 0.00 |
| 6 | x36 x37 x_46 + x_30 x59 x66 x68 + x78 x_92 x100 + x32 x_76 x_90 + x5 x_44 | 0.00 | 0.00 |
| 7 | x_30 x37 x_54 x68 + x17 x36 x_46 x78 + x_30 x_51 x64 x66 + x_44 x50 + x_41 | 0.00 | 0.00 |
| 8 | x14 x36 x66 x_71 + x_41 + x37 x68 x_70 + x37 x_44 + x7 x36 x55 x100 | 0.00 | 0.00 |
| 9 | x_30 x37 x_54 x68 + x36 x_46 x55 + x36 x55 x_80 + x_7 x_11 x_19 + x28 x50 x_60 + x_41 | 0.00 | 0.00 |
| 10 | x_15 x68 x_71 x83 + x_22 x37 x84 + x_18 x_76 x_90 + x36 x_48 x55 + x24 x31 x48 x51 + x_41 | 0.00 | 0.00 |
| Mean | | 0.00 | 0.00 |

## III. NDT APPLICATION

We investigate the potential of U-BRAIN algorithm for NDT. In order to provide a proof of concept, we use it as a rule-based clustering method to determine and classify the signal signatures of flaw characteristics such as size, depth, layer or angle by using an Eddy Current (EC) technique [27].

### A. Eddy Current Inspection

EC inspection is one of several NDT methods that use the electromagnetism principle as the basis for conducting examinations. Eddy currents are created through the process of electromagnetic induction. One of the major advantages of EC as an NDT tool is the variety of inspections and measurements that can be performed. ECs can be used for crack detection, material thickness measurements, coating thickness measurements, conductivity measurements, material identification, heat damage detection, damage depth determination. Furthermore EC is sensitive to small cracks, the inspection gives immediate results, the equipment is very portable, method can be used for much more than flaw detection, the test probe does not need to contact the part, is able to inspect complex shapes and sizes of materials. Nevertheless, a visual interpretation is usually used to analyze the data. Then, the results are influenced by subjectivity of human personnel. A more accurate data analysis can be obtained by solving complex multi-parametric partial differential equations. So, defect classification is generally carry out by signatures of the signal in the impedance plane, in the Fourier transform [27] or in principal component analysis (PCA) [28].

In this approach, the presence of damage is characterized by the changes in the signature of the resultant signal that propagates through the structure. For EC testing system, the response output signal is influenced from the material parameters. Let's note that the existence of defects in a material in the most of interesting cases lead to a significant alteration of its electrical characteristics. So, changing material parameters corresponds to a particular output signal that is characterized by a specific frequency spectrum.

In order to characterize the defect, the output signal is used as input to machine learning based classifiers. Most of the information in a signal is carried by its transient phenomena and its irregular structures. In such cases it is preferable to decompose the signal into elementary building blocks that are well localized in both time and frequency. This alternative can be achieved by using the Short Time Fourier transform (STFT) [29] and the Wavelet transform (WT) [30]. In this way, it is possible to define the local irregularity of a signal as special signature as input to an machine learning algorithm.

### B. Sample data

The data used in the study refers to a subset of a database with EC signal samples for aircraft structures [31]. The overall database is divided in 4 parts. The first contains 240 records acquired on an aluminum sample with notches of width 0.3 mm, depth 0.4, 0.7, 1, and 1.5 mm perpendicular, depth 0.4, 0.7, 1, and 1.5 mm with an angle of 30 degrees, 0.7, 1 and 1.5 mm with an angle of 60 degrees and 1.5 mm with an angle of 45 degrees. The second refers to 150 records, notches of width 0.2 mm, depth 1, 3 and 5 mm, both perpendicular and 45 degrees orientation of a stainless steel structure. Third database refers to two-layer aluminum aircraft structure with rivets, two notches below the rivets in the first layer (width 0.2 mm, length 2.5 mm, angle 90 degrees and 30 degrees) and two in the second layer (width 0.2 mm, length 2.5 mm and 5 mm, angle 90 degrees), two defect-free rivets. The fourth sample refers to four-layer aluminum structure (layer thickness 2,5 mm) with rivets containing 4 notches (width 0.2 mm, length 2.5 mm, angle 90 deg ) below the rivets in the first, second, third or fourth layer, four defect-free rivets.

### C. Pre-Processing

To extract and clustering features from EC signal signatures we use the Fourier transform of the two set of samples acquired on the aluminum structure. The first set refers to the notch perpendicular of width 0.3 mm, depth 1.5 mm. The second refers to the notch oblique of width 0.3 mm, depth 1.5 mm and angle of 60 degrees.

We made use of the Matlab program to perform spectrum analysis of the EC signals. The dataset contains 4096 samples, an sampling frequency of 10KHz and two canals for each acquired measure. The bandwidth is divided in 25 classes equally spaced. The first and last classes refers to frequency range starting from 0Hz and 9.7 KHz respectively. For each frequency classes the minimum, maximum, average and median of the FFT module are take in account to divide the overall range of the averages in 16 sub ranges. The thresholds of these ones ranges are chosen by considering the median value. Each sub range is codified by 4 bits in order to have 16 different levels representing the average value of FFT module in each frequency range.

*D. Rule-based Classifier*

To set up the U-BRAIN rule we use two set of data of 40 elements, each one of 100 variables, forming the positive and negative instances required to train the system. The classification performance of U-BRAIN is evaluated using a ten-fold cross-validation method. The results are reported in Table I. In the Table the underscore sign means a literal in negated form [13]. The algorithm has been executed on INTEL XEON E7xxx and E5xxx processor family with 24 cores on Linux-based cluster. The execution time of all the experiments was of few seconds.

## IV. CONCLUSIONS

In this paper we have discussed the use of a novel learning algorithm, named U-BRAIN, for the diagnosis of aerospace structure defects. The Soft Computing approach allows advanced multi-parameter data processing in composite materials testing. The HPC parallel implementation overcomes the limits due to the great amount of data and the complexity of data processing. Preliminary results on determining and classifying the signatures of flaw characteristics by using an eddy current technique show a surprising low error, zero, a result not obvious a priori, and confirm the U-BRAIN effectiveness in data analysis.

ACKNOWLEDGMENTS

This work has been supported in part by Distretto Aerospaziale della Campania (DAC) in the framework of the CERVIA project - PON03PE_00124_1.

APPENDIX

The following is the U-BRAIN algorithm schema. The symbol $S_{ij}$ refers to a set of constraints, while $R_{ij}$, $R_i$, $R$ are probability distributions (relevances).

1. Initialize f = Ø
2. While ∃ positive instances
2.1. Uncertainty Reduction
2.2. Repetition Deletion
2.3. Initialize term = Ø
2.4. Build $S_{ij}$ sets
2.5. While(∃ elements in $S_{ij}$)
2.5.1. Compute the $R_{ij}$ relevances
2.5.2. Compute the $R_i$ relevances
2.5.3. Compute the R relevances
2.5.4. Choose Literal
2.5.5. Update term
2.5.6. Update $S_{ij}$ sets
2.6. Add term to f
2.7. Update positive instances
2.8. Update negative instances
2.9. Check consistency